# PRIME: A CyberGIS Platform for Resilience Inference Measurement and Enhancement

Debayan Mandal[1], Dr. Lei Zou[1, *], Rohan Singh Wilkho[2], Joynal Abedin[1], Bing Zhou[1], Dr. Heng Cai[3], Dr. Furqan Baig[4], Dr. Nasir Gharaibeh[2], Dr. Nina Lam[5]

[1] Geospatial Exploration and Resolution (GEAR) Lab, Department of Geography, College of Arts & Sciences, Texas A&M University
[2] Zachry Department of Civil and Environmental Engineering, College of Engineering, Texas A&M University
[3] GIScience for Resilience (GIResilience) Lab, Department of Geography, College of Arts & Sciences, Texas A&M University
[4] CyberGIS Center for Advanced Digital & Spatial Studies, University of Illinois at Urbana–Champaign
[5] Department of Environmental Sciences, College of the Coast & Environment, Louisiana State University

* Corresponding author: lzou@tamu.edu

**Abstract:**

In an era of increased climatic disasters, there is an urgent need to develop reliable frameworks and tools for evaluating and improving community resilience to climatic hazards at multiple geographical and temporal scales. Defining and quantifying resilience in the social domain is relatively subjective due to the intricate interplay of socioeconomic factors with disaster resilience. Meanwhile, there is a lack of computationally rigorous, user-friendly tools that can support customized resilience assessment considering local conditions. This study aims to address these gaps through the power of CyberGIS with three objectives: 1) To develop an empirically validated disaster resilience model - Customized Resilience Inference Measurement (CRIM), designed for multi-scale community resilience assessment and influential socioeconomic factors identification; 2) To implement a Platform for Resilience Inference Measurement and Enhancement (PRIME) module in the CyberGISX platform backed by high-performance computing, enabling users to apply CRIM to compute and visualize disaster resilience; 3) To demonstrate the utility of PRIME through a representative study evaluating county-level community resilience to natural hazards in the United States. CRIM generates vulnerability, adaptability, and overall resilience scores derived from empirical parameters—hazard threat, damage, and recovery. Computationally intensive Machine Learning (ML) methods are employed to explain the intricate relationships between these scores and socioeconomic driving factors. PRIME provides a web-based notebook interface guiding users to select study areas, configure parameters, calculate and geo-visualize resilience scores, and interpret socioeconomic factors shaping resilience capacities. A representative study showcases the efficiency of the platform while explaining how the visual results obtained may be interpreted. The essence of this work lies in its comprehensive architecture that encapsulates the requisite data, analytical and geo-visualization functions, and

ML models for resilience assessment. This setup provides a foundation for assessing resilience and strategizing enhancement interventions.

## 1. Introduction

Under the threat of escalating natural hazards, the imperative for effective resilience assessment and improvement in the community setting is more critical than ever. As emphasized by the Intergovernmental Panel on Climate Change (IPCC, 2023), disaster resilience is defined as the capacity of social, economic, and ecosystems to cope with hazardous events, responding in ways that maintain their essential function, identity, and structure, while also maintaining the capacity for adaptation, learning, and transformation. Assessing and achieving community resilience to natural hazards is difficult due to several challenges. Resilience is a cross-disciplinary concept encompassing engineering, social, and environmental sciences. Although there are several physically validated studies to quantitatively define and enhance resilience in domains such as ecology and engineering, research in the social domain has remained mostly subjective due to the intricate nature of the relationships between socioeconomic factors and disaster resilience. Past efforts to develop models for evaluating disaster resilience have shown notable constraints, such as lacking empirical validation or designed for specific types of hazards or geographic regions, limiting their generalizability (Nelson et al., 2010a; Yusuf & Francisco, 2009). This calls for a customizable resilience assessment framework that is empirically validated and capable of effectively identifying influential driving factors to inform enhancement strategies.

Another challenge is the lack of versatile, accessible, and computationally powerful platforms to facilitate the implementation of customized resilience assessment by researchers and stakeholders (Li et al., 2015). Existing resilience assessment tools are often proprietary, requiring significant resources for implementation and ongoing maintenance. Consequently, stakeholders often resort to generic and opinion-based solutions, diminishing effectiveness. The lack of a customizable user-friendly platform also poses a barrier for domain experts lacking the technical skills to effectively engage with multi-source data for resilience research and practice. There is a need to develop a platform that can handle diverse data sets, has intuitive interfaces to help decision-makers access crucial data, and offers advanced tools for handling large datasets or looped models for evaluating and strengthening resilience.

Recent development of novel resilience assessment frameworks and technical advances, such as CyberGIS, a multidisciplinary field that combines cyberinfrastructure, geographic information systems (GIS), and spatial analysis (Wang et al., 2010), allow us to address these challenges head-on. On one hand, validated resilience assessment models, e.g., the Resilience Inference Measurement (RIM) framework, have been proposed and applied to evaluate community resilience to different types of natural hazards at varying scales (Cai et al., 2018). On the other hand, the ability of CyberGIS to seamlessly integrate data, methods, algorithms, and visualization into a unified platform for knowledge discovery and decision-making has led to its widespread adoption in addressing complex, large-scale, and multidisciplinary societal challenges (Yin et al., 2019; Wang 2010, 2013). Its effectiveness in handling big data and intensive computational tasks for diverse fields, such as public health (Shi & Wang, 2015), environmental management (Choi et al., 2021), and urban planning (Shirowzhan et al., 2020), validates its utility. The influx of big data in the social domain with respect to disaster resilience necessitates the application of CyberGIS to increase efficiency by providing a practical computational time of spatial analysis.

This study is poised to address the above-mentioned challenges and explore the potential of cyberinfrastructure and advanced resilience assessment frameworks in disaster resilience research. The objectives are three-fold: (1) To develop an empirically validated disaster resilience model - Customized Resilience Inference Measurement (CRIM), designed for multi-scale community resilience assessment and the identification of influential socioeconomic factors; (2) To implement a Platform for Resilience Inference Measurement and Enhancement (PRIME) module in the CyberGISX platform backed by high-performance computing, enabling users to apply CRIM to compute and visualize disaster resilience; (3) To demonstrate the utility of the CyberGIS platform through a representative study evaluating county-level community resilience to natural hazards in the United States. PRIME provides a promising avenue for addressing the challenges of disaster resilience, in terms of ease of accessibility and reproducibility. It also promises better outreach and future planning for including computationally intensive methodological improvements.

The article is organized as follows. Section 2 reviews previous efforts that have defined resilience and identified socioeconomic indicators to quantify resilience. Section 3 introduces the CRIM model and explains the enhancements made compared to existing models. In Section 4, we introduce the interface of the CyberGIS tool, PRIME, and how it functions. Section 5 delves into the experimental design of the current platform. Section 6 shows how our findings can be used in a real-world setting. Finally, Section 7 concludes our study and suggests areas for implementation of the PRIME.

## 2. Literature Review

### 2.1 Resilience Definition

The definition of resilience, although a deep-rooted concept in social-ecological literature, has remained inconsistent, with interpretations varying from capacity, process, and outcome to a combination of these aspects (Cai et al., 2018). Meanwhile, the definition of resilience often depends on the field of application, introducing further complexity to the underlying task of understanding and measuring it. Resilience's relationship with vulnerability and adaptability has also been a key focus in the discourse among researchers (National Research Council, 2012).

Various researchers have proposed different definitions over the years. Holling (1996) introduced the concept of engineering resilience as the speed at which a system recovers to its equilibrium after a disturbance, and ecological resilience as the extent of disturbance a system can withstand before altering its structural components and behavioral controls. Further expanding on the concept, Turner et al. (2003) characterized resilience as a system's capacity to revert to a reference state and maintain its structural and functional attributes post-disturbance. In the social science domain, Cutter et al. (2008) defined resilience as the inherent and adaptive capacities of a social system to respond, recover, and learn from disasters. Yin et al. (2022) added to the ongoing dialogue by perceiving resilience as the establishment of the capacity to effectively rebuild after a tragedy. The repeated attempts at defining resilience reflect its multi-dimensional and context-dependent nature.

Apart from individual studies defining resilience, there were thorough reviews conducted for this purpose. In the National Research Council's report in 2012 (National Research Council, 2012), an extensive review was conducted to standardize its articulation - Resilience is the ability to prepare and plan for, absorb, recover from, and more successfully adapt to adverse events. Subsequently, in their broad systematic review of 174 disaster resilience studies, Cai et al. (2018) summarized the most frequently used keywords to define disaster resilience over several disciplines of resilience research within 2005 to 2018. It was found that regardless of the discipline or disaster type, 'system', 'ability', 'capacity', and 'recover' remained the most frequently used words in defining resilience.

## 2.2 Resilience Assessment Frameworks

In addition to defining disaster resilience, numerous frameworks have been proposed to quantify resilience. Many notable existing resilience and relevant indices, such as the social vulnerability index (Cutter et al., 2003) and the community resilience system (CRS) (Plodinec, 2013), provide different methods of approaching resilience measurement. In general, resilience indexing methods can broadly be classified into context-specific index, comprehensive index, and empirical index. Table 1 summarizes commonly used quantification approaches of resilience and relevant concepts with their strengths and limitations.

Table 1. Different Disaster Resilience Assessment Models

| Model/Study | Methodology | Strengths | Limitations | References |
|---|---|---|---|---|
| **Context Specific Index** | | | | |
| Social Vulnerability Index (SoVI) to environmental hazards | Weighted aggregation of socioeconomic variables through statistical methods | Ease of index derivation | Lack of empirical validation such as correlating the index with disaster declarations | (Cutter et al., 2003) |
| Vulnerability to Climate Change | Weighted aggregation of exposure, sensitivity, and adaptive capacity | Straightforward comparison among regions | Subjective nature of weight assignments; no empirical validation | (Yusuf & Francisco, 2009) |
| **Comprehensive Index** | | | | |

| Vulnerability and Adaptive Capacity of Australian Rural Communities | Creation of a weighted adaptive capacity index | Provides comprehensive localized adaptive capacity (social, economic, and environmental) | Lack of empirical validation; inapplicable beyond Australian rural contexts | (Nelson et al., 2010b) |
|---|---|---|---|---|
| Baseline Resilience Indicator for Communities (BRIC) | Arithmetic mean of five subindexes (i.e., social, economic, institutional, infrastructure, community) | Consideration of multiple aspects of Resilience | Lack of empirical validation for the composite indicator | (Cutter et al., 2010) |
| Indicators of Capacities for Community Resilience | Correlation analysis and additive index | Comprehensive list of variables | Lack of empirical validation for the composite indicator | (Sherrieb et al., 2010) |
| **Empirical Index** | | | | |
| Resilience Inference Measurement (RIM) | Calculation of empirical resilience index based on hazard threat, damage, and recovery | Empirically validated index | Implementation difficulties due to need for extensive data collection and technical expertise | (Lam et al., 2015) |

Context-specific index method assesses resilience to specific contexts of the community, such as social (Cutter et al., 2003), economic (Yusuf & Francisco, 2009), etc. It also recognizes that resilience factors can vary depending on the spatiotemporal specifics of the region or community. The selection of factors and their interpretation is context-specific, and they are assigned weights using either expert judgements or techniques like Analytic Hierarchy Process (AHP), Principal Components Analysis (PCA), etc. Consequently, the additive weighting mechanism is used to combine these factors into a single composite index. The method is straightforward and allows for easy comparability. However, the context-specific nature of these indices means that their broad applicability might be limited. Some of the commonly used context-specific indexes lack empirical validation - a disconnection between its theoretical concept and practical application (Spielman et al., 2020).

Building on the context-specific approach, the comprehensive index method integrates a wider range of factors. For instance, based upon the social vulnerability index proposed in 2003, Cutter et al. (2010) adopted a more comprehensive approach to incorporate the economic, environmental, institutional, community, and infrastructure domains into the equation in addition to the social to calculate the Baseline Resilience Indicator for Communities (BRIC). Nelson et al. (2010a) took into account the social, economic, environmental aspects in the quantification of resilience through vulnerability and adaptive capacity. They applied statistical models similar to models used in the context-specific index in aggregating the indices for the different dimensions. This method represents an advancement in its holistic approach, acknowledging the complexity and interrelated nature of resilience factors. It balances offering a comprehensive overview and the risk of becoming too generalized. However, these indexes have similar limitations like context-specific methods due to the lack of validation.

The empirical index approach follows the comprehensive index as an improvement in disaster resilience indexing. This method is based on observed data moving beyond theoretical or assumed relationships between indicators and resilience, validating these relationships through empirical evidence with advanced statistical techniques, e.g., regression analysis and machine learning algorithms. This approach enhances the reliability and accuracy of resilience measurements by grounding them in empirical evidence. Due to their data-driven statistical approaches, they are often better suited to capture the evolving nature of resilience through predictive modelling. The Resilience Inference Measurement (RIM) model (Lam et al., 2015) extends research in that direction. It evaluates resilience based on communities' real impacts and recoveries under disasters and leverages machine learning methods to identify influential factors shaping empirically observed resilience capacity. In this way, the identified resilience indicators and indexes have been validated through ground evidence and are more reliable in predicting resilience changes or in other regions lacking historical disaster information.

In a nutshell, resilience research, historically, dealt with the issue of no universally accepted definition and quantification. Partially due to this, existing resilience indices developed by different investigations may yield discrepant results, and most of them lack empirical validation, making it difficult to select suitable resilience indices in practice and decision-making.

## 3. Customized Resilience Inference Measurement (CRIM) Model

The CRIM model proposed in this research is based on the RIM model, which was originally developed to quantify county resilience to coastal hazards (Lam et al. 2015). This section first introduces the RIM model and then elaborates on the improvements made in the CRIM model.

### 3.1. Resilience Inference Measurement (RIM) Model

The RIM model's concept of resilience aligns with the National Research Council's report (NRC 2012) which considers resilience to comprise both vulnerability and adaptability (Figure 1). Vulnerability and adaptability are assessed through three empirical factors for disasters, i.e., threat, damage, and recovery. The experimental design of these three factors may differ based on the hazards targeted. Theoretically, threat was the incidence/intensity of the hazard, damage was the actual impacts caused by the hazard, e.g., infrastructure malfunction, monetary loss, or fatality due to the hazard, and recovery was the return of the infrastructure functions, population, or economic growth (Mihunov et al., 2019). The model defines vulnerability as a community's susceptibility to the erratic fluctuations of natural disasters and quantifies it as the slope of the threat-damage curve. Adaptability is portrayed as a community's capacity to recuperate from a disaster and revert to normal activity levels and estimated as the slope of the damage-recovery curve.

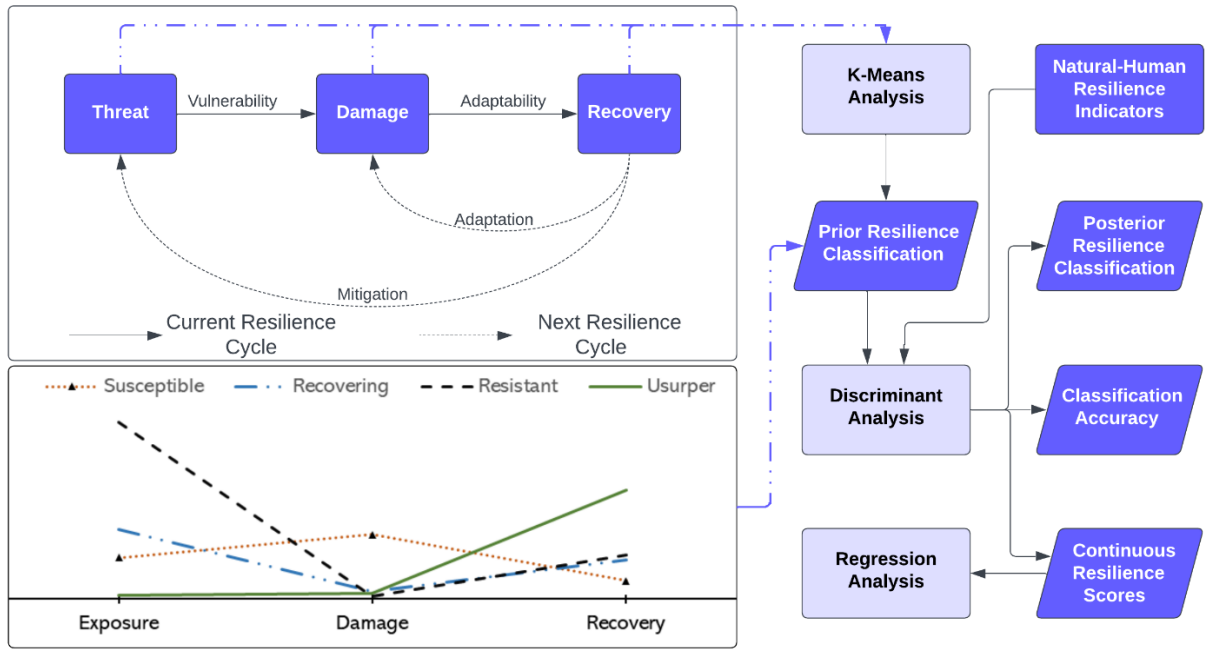

*Figure 1. The concept of resilience defined by the Resilience Inference Measurement (RIM) framework.*

Table 2 illustrates a list of studies that have employed the RIM model (for different disaster types and study areas) and their definitions of the three empirical disaster factors: threat, damage, and recovery. They have consistently followed three steps (Figure 1). First, relevant disaster factors were computed for selected study areas. Second, to perform preliminary resilience categorization, clustering methods such as the k-means clustering were used to classify communities into one of the four clusters, inspired by ecological resilience studies (Batista & Platt, 2003; Bellingham et al., 1995): (1) Susceptible - High vulnerability and Low adaptability, (2) Recovering - Medium vulnerability and Medium adaptability, (3) Resistant - Low vulnerability and Medium adaptability, and (4) Usurper - Low vulnerability and High adaptability. Third, machine learning algorithms (e.g., discriminant analysis) were leveraged to delineate the association between the empirical resilience index and socio-economic determinants.

Table 2. Resilience Inference Measurement Model Studies

| Case Study | Disaster type | Study Area | Time Period | Threat | Damage | Recovery |
|---|---|---|---|---|---|---|
| Lam et al., 2015 | Hurricanes | 25 countries in the Caribbean region | 1987 to 2012 | Hurricane Recurrence | Monetary loss per capita | Population change rate (1987-2012) |
| K. Li et al., 2015 | 18 different natural hazards (thunderstorms, hurricanes, floods, wildfires, and others.) | Counties and States in the U.S. | 2000 to 2010 | Weighted hazard frequency | Property and Crop damage per capita | Population change rate (2000-2010) |
| Lam et al., 2016 | Coastal hazards (coastal flooding and storm surge), flood, hurricane, thunderstorm, and tornado. | 52 U.S. counties along the northern Gulf of Mexico | 1998 to 2008 | Weighted hazard frequency | Property and Crop damage per capita | Population change rate (1998-2008) |
| Cai et al., 2016 | Storm surge, flood, hurricane, tropical storm, and tornado | 2086 block groups in the Lower Mississippi River Basin | 2000 to 2010 | Weighted duration per hazard type | Property damage per capita | Population change rate (2000-2010) |
| X. Li et al., 2016 | 2008 Wenchuan Earthquake | Sichuan, Shaanxi, and Gansu Provinces, China | 2002 to 2011 | earthquake Intensity of the earthquake | Direct economic losses per capita | Population change rate (2002-2011) |
| Mihunov et al., 2018 | Droughts | 503 counties in Arkansas, Louisiana, New Mexico, Oklahoma, and Texas | 2000 to 2015 | Number of weeks of severe drought duration | Total amount of insured crop losses per capita | Population change rate (2000-2015) |
| Lam et al., 2018 | Storm surges, floods, hurricanes, tropical storms and tornados | 2086 block groups in the Lower Mississippi River Basin | 2000 to 2010 | Weighted duration per hazard type | Property damage per capita | Population change rate (2000-2010) |
| Mihunov et al., 2019 | Droughts | 503 counties in Arkansas, Louisiana, New Mexico, Oklahoma, and Texas | 2000 to 2015 | Number of weeks of severe drought duration | Total amount of insured crop losses per capita | Population change rate (2000-2015) |
| K. Wang et al., 2021 | 2012 Hurricane Isaac | 146 affected counties | 2011 to 2013 | Average wind speed | Property and Crop damage per capita | Population change rate (2011-2013) |

However, these studies are subject to several limitations. For instance, when testing the association between resilience scores and socio-economic characteristics, there is a temporal inconsistency in the socio-economic variables, where all socio-economic variables are not of the same year, possibly due to data unavailability. For instance, Mihunov et al., (2019), in their study period of 2000-2015, have used a total of 52 variables, including 19 variables from 1997, 32 variables from 2000, and 1 variable from 2014. This introduces an uncertainty into the framework. The socioeconomic variables are assumed static but are inherently dynamic. Thus, the variables used to characterize the affected community may have changed during the period of study. Additionally, these studies evaluate the model performance based on the same dataset used to train the algorithms; whereas machine learning theories suggest that algorithms must be evaluated on a held-out testing set. Otherwise, as the model already "sees" the data albeit in different folds, even cross-validation is unable to prevent overfitting as the model might just be recalling data rather than predicting it (Kaufman et al., 2012). Furthermore, the current ecological studies-based classifications of resilience tend to neglect different categorizations like High Vulnerability with Low and Average Adaptability; Average Vulnerability with Low and High Adaptability; and Low Vulnerability with Low Adaptability. The final recurring issue pertains to the use of different machine learning modules to predict the dependent variable and quantify their sensitivity. For instance, regression analysis has been frequently used to delineate the relationships between socioeconomic variables (independent variables) with the resilience classification. This method might not capture non-linear relationships. To overcome these limitations, our study presents CRIM, an enhanced, customizable framework of the RIM model.

### 3.2. Enhanced Customizable RIM (CRIM) Framework

In formulating the CRIM framework, we have revised the fundamental RIM framework to address its limitations. The CRIM workflow and the changes compared to the RIM model are displayed in Figure 2.

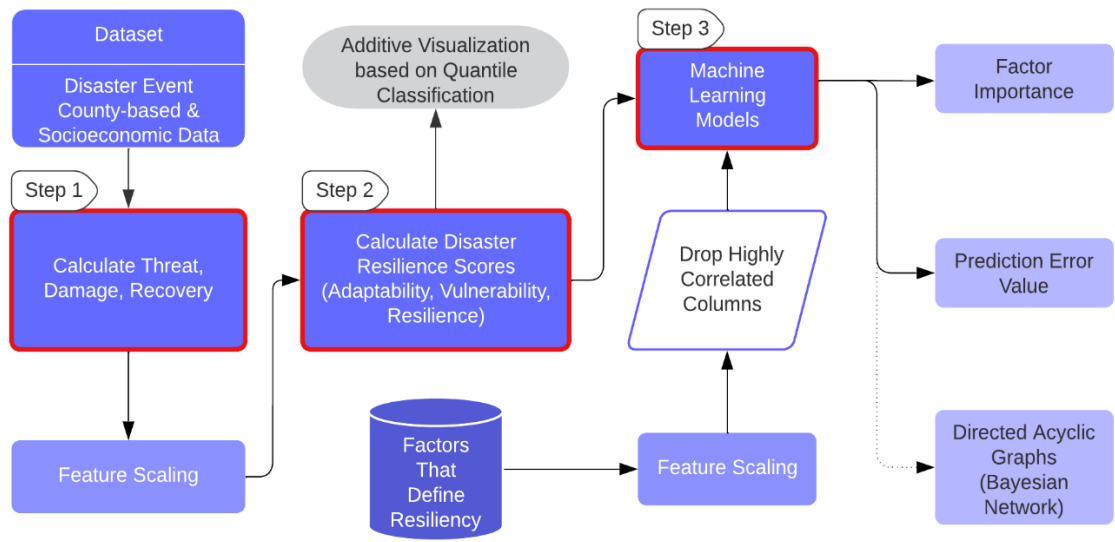

*Figure 2. Workflow for the CRIM Model*

The CRIM framework first computes the three empirical parameters: threat, damage, and recovery. A unique update of the CRIM model compared to the RIM model is incorporating a comprehensive weighting mechanism to quantify the overall threat a community suffered from different types of hazards using Equation 1 (FEMA, 2023). Equation 1 considers three key factors of hazard threat: the duration of the hazard event, the likelihood of hazard occurrence, and the weight assigned to each type of hazard event. We calculate the cumulative threat values for each county by summing up these values for each hazard event in each year.

$$Threat_{county} = \sum_{event} (Duration_{days} * Likelihood_{hazard} * Weightage_{hazard})$$

(Equation 1)

In Equation 1, duration refers to the length of the historic hazard events as days. This variable recognizes that the impact of a hazard event typically increases with its duration. The second component is the likelihood, which quantifies the probability frequency distribution of hazard type per day (Equation 2). This factor acknowledges the inherent uncertainty in the occurrence of hazard events. The third component is the weight for each hazard type, which is depicted as the mean damage caused per day per capita (Equation 3). This factor recognizes that different hazard types can have varied potential impacts based on their nature. For example, tornadoes might occur more frequently but cause less average damage than earthquakes, so the threat of one tornado should be lower than the threat of one earthquake.

$$Likelihood_{hazard} = Count\ /\ Total\ Days \qquad \text{(Equation 2)}$$

$$Weightage_{hazard} = Mean\ Damage\ per\ day\ per\ capita \qquad \text{(Equation 3)}$$

Continuing Step 1 is calculating damage, the second empirical parameter, using Equation 4. Disaster damage is multidimensional and includes different types, e.g., crop and property damage. For each hazard event, we derive a per capita estimate of the damage by dividing the sum of damages considered in the specific case study by the pre-event initial population ($Population_{Initial}$). The final empirical parameter, disaster recovery, can be determined as the recovery rate of population, employment, economy, or other indicators after disasters. Population recovery is the most used recovery indicator (Equation 5). In the equation, $Population_{Final}$ is the population count recorded in the year following the disaster. A higher recovery rate for a community signifies an increase in population after the disaster events, indicating successful society rebuilding efforts.

$$Damage_{county} = \sum_{event} \left( \frac{Damage_{Total}}{Population_{Initial}} \right) \qquad \text{(Equation 4)}$$

$$Recovery\ Rate_{county} = \frac{(Population_{Final} - Population_{Initial})}{Population_{Initial}} \qquad \text{(Equation 5)}$$

The second step of the CRIM workflow is deriving resilience scores. Compared to the RIM model, CRIM substitutes the frequently used k-means clustering approach with a composite scoring system for the indices to enhance model precision. The three empirical parameters (threat, damage, and recovery) are normalized using a min-max scaler (Pedregosa et al., 2011) and used to calculate the two relations: vulnerability, and adaptability. Min-max scaling is particularly beneficial when the data does not follow a Gaussian distribution, which was common with the empirical parameters. This method preserves the shape of the original distribution and the importance of outliers, thus making it a fitting choice for non-parametric data. Vulnerability is computed as the difference between the normalized values of damage and threat (Equation 6).

The underlying principle contends that a community enduring equivalent destruction from infrequent disasters would be perceived as more vulnerable compared to a community exposed to more recurrent disasters. The Adaptability Score is computed as the difference between the normalized values of recovery and damage (Equation 7). In both cases, the difference is obtained after normalization as both the minuend and subtrahends are of different units. Lastly, the empirical Resilience score is a comprehensive measure and interpreted as the relation between the adaptability and vulnerability of a community (Equation 8). Equation 8 implies that increasing a community's adaptability or reducing its vulnerability would improve the community's overall resilience capacity to disasters. These indices can be computed yearly or over a longer period (e.g., 10 or 20 years) to offer insights into the dynamics of disaster impact and recovery.

$$Vulnerability = Damage_{Normalized} - Threat_{Normalized} \qquad \text{(Equation 6)}$$

$$Adaptability = Recovery_{Normalized} - Damage_{Normalized} \qquad \text{(Equation 7)}$$

$$Resilience = Adaptability - Vulnerability \qquad \text{(Equation 8)}$$

Step 3 identifies the socioeconomic indicators influencing communities' resilience capacities. CRIM associates resilience indices per year for communities experiencing natural hazards with the previous year's socioeconomic variables, which accurately represent the pre-disaster conditions, through ML algorithms. This approach can resolve the inconsistency issue in the socio-economic variables in the original RIM model. In preparation for the third step, CRIM normalizes the independent socioeconomic parameters to ensure that features with larger numeric ranges do not dominate over the ones with smaller numeric ranges. As multiple socioeconomic factors did not follow the Gaussian distribution, the normalization technique used was min-max scaler. Alongside that, some socioeconomic parameters tend to be collinear which violates the assumption of independent factors in some ML models. Hence, we remove the collinear variables, retaining only one factor from each set of correlated factors.

These normalized and cleaned socioeconomic variables were used as independent variables in training ML models. This methodology is grounded on the assumption that a community's vulnerability and adaptability to disasters are fundamentally dependent upon its socioeconomic characteristics (Maclean et al. 2014). Variations in these socioeconomic indicators would account for differences in disaster resilience capacities, providing a potential predictive framework. The use of ML models provides final validation metrics and associations and informs causations of the three scores with the socioeconomic variables. CRIM validates the associations by calculating absolute error over a distinctly held-out test dataset, which is distinct from the original RIM model. Relationships derived from these associations and potential causations could help in evidence-based policymaking and inspire future research, such as comparative analysis and predictive studies.

## 4. The Platform for Resilience Inference Measurement and Enhancement (PRIME)

### 4.1. User Interface

The CRIM framework is incorporated in PRIME, a CyberGIS platform hosted in CyberGISX (Wang, 2010). CyberGISX is a state-of-the-art cyberinfrastructure developed and maintained by the University of Illinois Urbana-Champaign (https://cybergisxhub.cigi.illinois.edu/). It combines advanced computing and geospatial technologies, and supports a wide range of applications, from environmental modeling to urban planning (Wang, 2016). By hosting the CRIM model on this platform, we ensure reproducible and accessible resilience assessment tools at varying spatial and temporal scales.

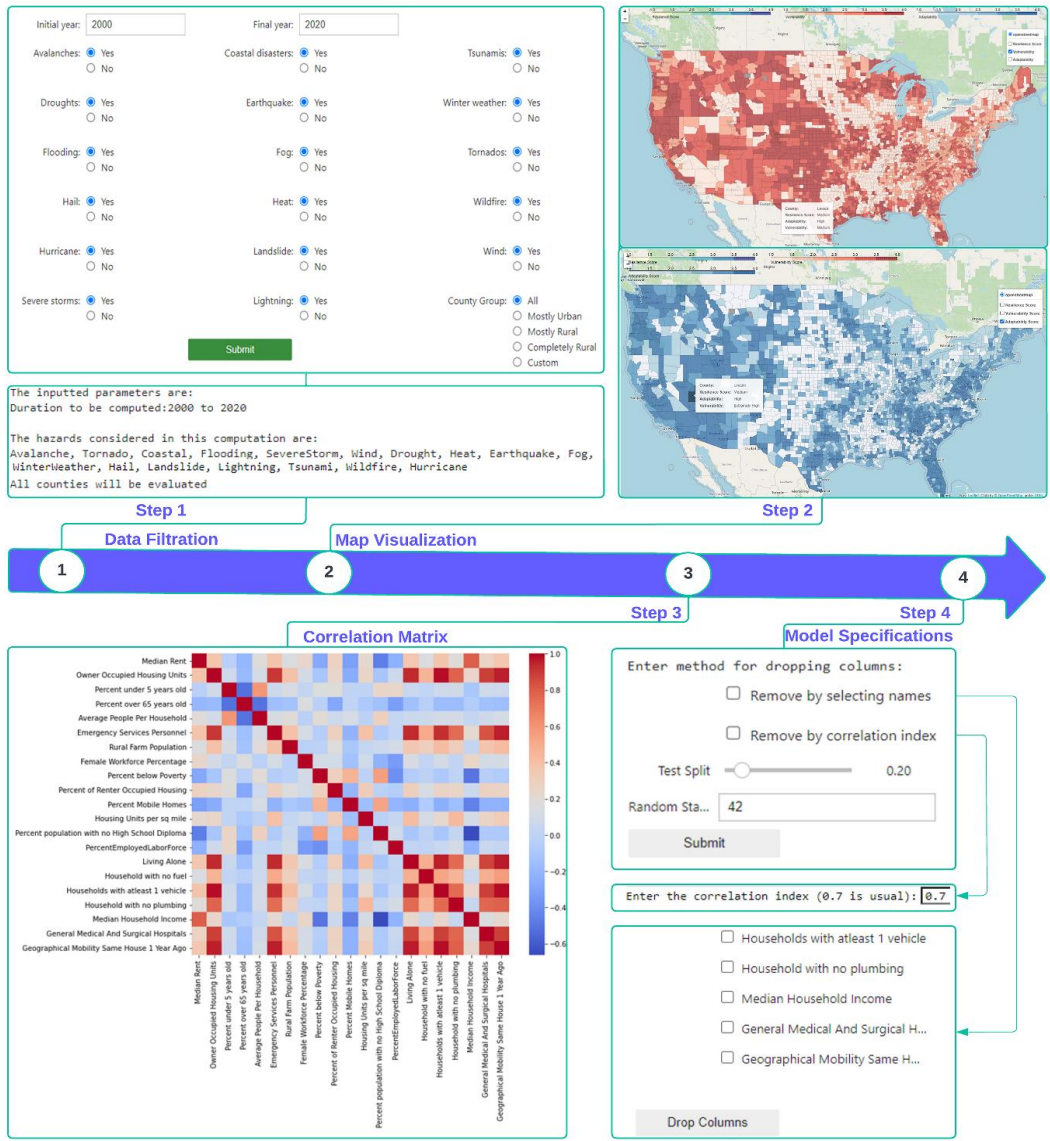

*Figure 3. User interfaces and workflows for (1) defining parameters to filter disaster dataset, (2) interactive visualizations of resilience scores, (3) visualizing correlation matrix, and (4) elimination options for correlated variables and input for train-test split.*

PRIME offers four user interfaces that need to be executed sequentially as it appears through the workflow (Figure 3). The first interface (Step 1 in Figure 3) enables the selection of parameters for filtering the disaster dataset as per the study requirements to compute resilience scores. Alongside, the framework incorporates a quantile classification function that transforms the continuous resilience scores into four categories, for exploratory data analysis. The second interface visualizes these empirical resilience, vulnerability, and adaptability classes, in different choropleth layers, across different counties via an interactive map (Step 2 in Figure 3). These layers are initially set to be hidden and can be made visible through the layer control. When a county on the map is hovered over – the county name, resilience, adaptability, and vulnerability scores are displayed in a tooltip.

The third interface offers a correlation test among the socio-economic variables and an option for correlated variable removal. This feature caters to models that require independent variables to be non-correlated. To guide variable removal, a correlation matrix is printed in the workspace (Step 3 in Figure 3). In the final interface (Step 4 in Figure 3), users can manually select the socioeconomic variables to eliminate or apply a correlation index magnitude method. The correlation index magnitude method, irrespective of positive or negative correlation, will remove highly correlated socioeconomic variables, retaining the first one. Alongside, the users can specify a ratio for train-test data splitting and set a random seed for the machine learning model.

### 4.2. Available Machine Learning Models

PRIME offers a selection of machine learning models to implement CRIM for resilience assessment and modeling, covering both white-box (fully interpretable) and grey-box (partially interpretable) models (Table 3). These models are applied to associate the resilience scores with socioeconomic factors and develop a validated predictive model. The choice of the most precise model for such predictive tasks is left to the users. To provide explainability, the framework also generates graphs depicting the relationship between the socio-economic variables and resilience scores based on selected ML models using coefficients, importance scores, or directed arcs. The white-box models generate graphs depicting positive and negative coefficients or causal arcs for each independent variable, while grey-box models demonstrate specific feature importance. This visualization helps identify influential socioeconomic factors, assisting in evidence-based policy development.

Table 3: Machine Learning Models incorporated in PRIME.

| Model | Brief Description | Explainability offered by |
|---|---|---|
| Linear Regressor (WB) | Linear regression that best fits the data by minimizing the sum of squared errors. | Coefficients |
| Ridge Regressor (WB) | Linear regression that addresses multicollinearity | |
| Lasso Regressor (WB) | Linear regression for conditions when a few predictors have a strong impact. | |
| Polynomial Regressor (WB) | Capturing non-linear relationships by fitting a polynomial line (*n*th degree). | |
| Random Forest Regressor (GB) | Combines multiple decision trees to create a predictive model. | Importance Score |
| XGBoost Regressor (GB) | Machine learning algorithm that is based on gradient boosting of decision trees. | |
| Bayesian Network (WB) | Graphical model that represents the probabilistic relationships among a set of variables, by depicting conditional dependencies through a directed acyclic graph. | Directed Arcs |

Abbreviations: WB – White Box Model; GB – Grey Box Model

The framework for each ML model follows the sequence of hyperparameter tuning, re-training tuned models and model evaluation. PRIME optimizes model performance through hyperparameter tuning, selecting models yielding the lowest MAE values in the cross-validation (Pedregosa et al., 2011) process. After determining the optimized hyperparameters, it initializes the models with these parameters and re-trains them on the original training set. Subsequently, it assesses the model performance by computing evaluation metrics on the held-out testing set, improving upon the erstwhile RIM model. The model performance will be assessed using Mean Squared Error (MSE), Root Mean Squared Error (RMSE), and Mean Absolute Error (MAE) metrics.

### 4.3. Embedded Data

The PRIME incorporates two primary datasets: hazard data from the Spatial Hazard Event and Losses Database for the United States (SHELDUS) (*Arizona State University Center for Emergency Management and Homeland Security*, 2023) and Census data (*U.S. Census Bureau, 2020*). The SHELDUS dataset provides valuable information on hazard events including location and time of occurrence, crop and property damages and fatalities per hazard event from 1960 to present. This encompasses a wide range of disasters such as avalanches, tornadoes, coastal hazards, flooding, severe storms, wind events, droughts, heatwaves, earthquakes, fog, winter weather, hailstorms, landslides, lightning, tsunamis, wildfires, and hurricanes. Crop damage refers to the harm inflicted on agricultural produce by the disaster, which can impact local economies, particularly in areas heavily dependent on agriculture. Property damage pertains to the destruction of infrastructure (homes, businesses, public facilities) which has direct and immediate impact on urban residents' living conditions and livelihoods.

The processed data available in PRIME consist of adjusted damage per capita (based upon monetary value to 2020) for every hazard event based on the county it occurred in, along with the date of occurrence. To make PRIME generalizable, we offer users the flexibility to add hazard information datasets pertinent to their research interests in the designated "data/disaster" folder on the server. We recommend that users incorporate damages that are more likely to be the result of the disaster type being studied. For example, drought resilience studies should primarily consider crop damage. The processed disaster dataset contains five columns delineating the disaster-related metrics: 'UniqueCode' representing the FIPS county code, 'Disaster' identifying the hazard type, 'Year' noting the occurrence year, 'DamageRIM' containing adjusted total damage per capita, and 'Duration (days)' storing the disaster's duration.

Second, the potential socio-economic variables from the Census data are incorporated and pre-stored in PRIME to identify the driving factors and estimate their influencing degrees. This data is available in the US Census Bureau for the year 2000 and the years 2010 – 2020. For the years where data is not directly available, we employed the robust Seasonal and Trend decomposition using the Loess (STL) method (Cleveland et al., 1990) to generate the necessary data points. As users engage with the CyberGISX, they will find all this data readily available across a significant temporal span (2000 – 2020). The choice of these parameters is based on the seminal works of Cutter et al., 2010, Lam et al., 2015, and Cai et al., 2018. Table 4 contains the default list of socio-economic parameters. Users may choose a subset from the list or add on to this list in the 'data/variables' folder on the server platform.

Table 4: Socio-economic factors embedded in PRIME (Source: US Census).

| Category | Indicators |
|---|---|
| Demographic | Percentage of population under 5 years old |
| | Percentage of population over 65 years old |
| | Average number of people per household |
| | Rural farm population |
| | Percentage of the workforce that is female |
| | Percent single households |
| | Percent of the population over 25 with no high school diploma |
| Economic | Median Rent |
| | Median Household Income |
| | Percent of the population living below poverty |

|  | Percent of the labor force that is employed |
| --- | --- |
| **Housing and Living Conditions** | Number of owner-occupied housing units |
|  | Percentage of renter-occupied housing |
|  | Percent of homes that are mobile homes |
|  | Number of houses per square mile |
|  | Households with at least one vehicle |
|  | Households with no fuel used |
|  | Households with no plumbing |
| **Community Resources** | Geographical Mobility (Same House 1 year ago) |
|  | Number of Hospitals |
|  | Emergency Services Personnel |

## 5. Representative Study

The representative study uses the CRIM framework in PRIME to measure vulnerability, adaptability, and resilience scores for every county in the US across all available disaster types from 2000 to 2020. We employ ML models that produce understandable and sufficiently accurate results. We subsequently discuss these results, focusing on their benefits, drawbacks, and interpretability. A litany of ML models is tested in the case study, including linear regression, ridge regression, polynomial regression, Bayesian network, random forest, and XGBoost. To remove collinearity amongst the independent variables, we omitted the socioeconomic variables having correlations exceeding 0.7. This resulted in removing variables such as General Medical and Surgical Hospitals, Median Household Income, Living Alone, Emergency Services Personnel, Households without Plumbing, Geographical Mobility of the Same House over 1 Year Ago, and Households with at least 1 Vehicle due to their high correlation with Owner-Occupied Housing Units.

Table 5 presents three performance metrics for all selected models corresponding to each resilience score. Low MSE and RMSE values suggest that the models' predictions closely match the actual values, indicating an excellent model fit. The MAE, which is not sensitive to outliers, provides an estimate of the typical prediction error. Lower MAE values suggest better-performing models. In general, all selected models exhibit reliable performance, making them suitable for decision-making in disaster management planning. We also found that the Ridge Regression and Linear Regression models yield identical results. This is because we removed collinear independent variables in the pre-processing stage, making the Ridge Regression's ability to handle multicollinearity redundant. Meanwhile, Random Forest models consistently outperformed XGBoost. Bayesian Network has also performed the best amongst all in terms of RMSE and MSE.

Therefore, the rest of this section focuses on interpreting the results from Linear Regression, Random Forest, and Bayesian Network.

Table 5. Machine Learning models performance

| Machine Learning Regressors | Mean Squared Error | Root Mean Squared Error | Mean Absolute Error |
|---|---|---|---|
| *Vulnerability Score* | | | |
| Linear Regression | 0.00016 | 0.01252 | 0.00176 |
| Ridge Regression | 0.00016 | 0.01252 | 0.00176 |
| Random Forest Regression | 0.00015 | 0.01241 | 0.00179 |
| XGBoost Regression | 0.00015 | 0.01250 | 0.00182 |
| Bayesian Network | **0.00013** | **0.01167** | **0.00168** |
| *Adaptability Score* | | | |
| Linear Regression | 0.00034 | 0.01845 | 0.00691 |
| Ridge Regression | 0.00034 | 0.01845 | 0.00691 |
| Random Forest Regression | 0.00031 | 0.01786 | **0.00579** |
| XGBoost Regression | 0.00034 | 0.01863 | 0.00651 |
| Bayesian Network | **0.00024** | **0.01553** | 0.00722 |
| *Resilience Score* | | | |
| Linear Regression | 0.00052 | 0.02278 | 0.00530 |
| Ridge Regression | 0.00052 | 0.02278 | 0.00529 |
| Random Forest Regression | 0.00050 | 0.02251 | **0.00466** |
| XGBoost Regression | 0.00050 | 0.02265 | 0.00483 |
| Bayesian Network | **0.00022** | **0.01514** | 0.00532 |

Figure 4 illustrate the spatial distribution of vulnerability, adaptability, and resilience scores across the United States at the county level for 2000 to 2020. Figure 4(a) portrays vulnerability, where shades of blue represent lower vulnerability, transitioning to red as vulnerability increases. The central regions (Colorado, Kansas, New Mexico, Oklahoma) show a higher vulnerability, contrasted by the Northeast (New Jersey, Detroit, District of Columbia) and some parts of West (Southern California sans coastal counties, Eastern Idaho) which demonstrate moderate to low vulnerability. Figure 4(b) depicts adaptability, where shades of blue represent higher adaptability, transitioning to red as adaptability decreases. Coastal areas, particularly on the eastern seaboard (Florida, North Carolina) and the west coast (California, Oregon, Washington), exhibit higher adaptability, whereas a big portion of the central U.S. (Northern Texas, Kansas, Nebraska, Eastern North Dakota) displays lower adaptability levels. Figure 4(c) showcases resilience with a disparate spatial pattern: pockets of high resilience are scattered predominantly in the Midwest (Utah, Western Colorado, Oregon) and some coastal areas (Eastern Florida, California, Oregon), while areas of low resilience appear to be dispersed across the Centre (Kansas, Nebraska), South (Northern Texas) and parts of the Northeast (Maine).

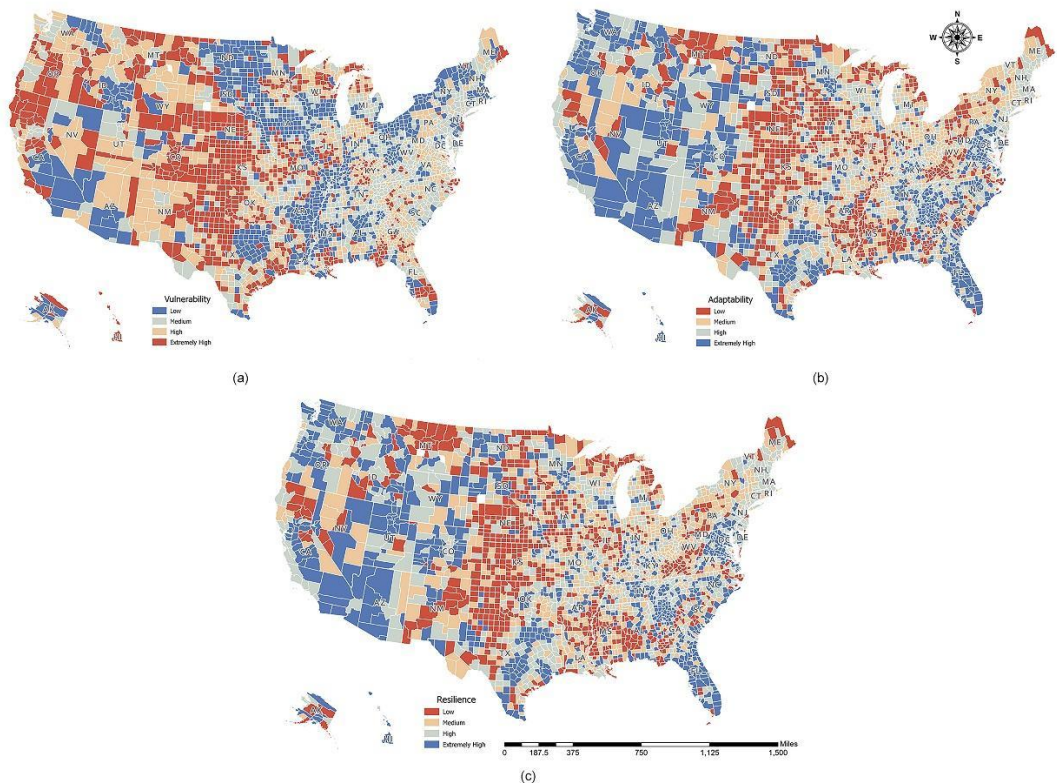

*Figure 4. County-level patterns of overall empirical vulnerability (a), adaptability (b), resilience (c) scores in the United States for 2000-2020*

Figure 5a, 5b and 5c display the coefficients and feature importance graphs generated by the Linear Regression and Random Forest models for the vulnerability, adaptability, and resilience scores, respectively. These are obtained separately in PRIME and consolidated for brevity. There are several inferences that can be drawn from these results. For concision, a few of these factors shall be discussed.

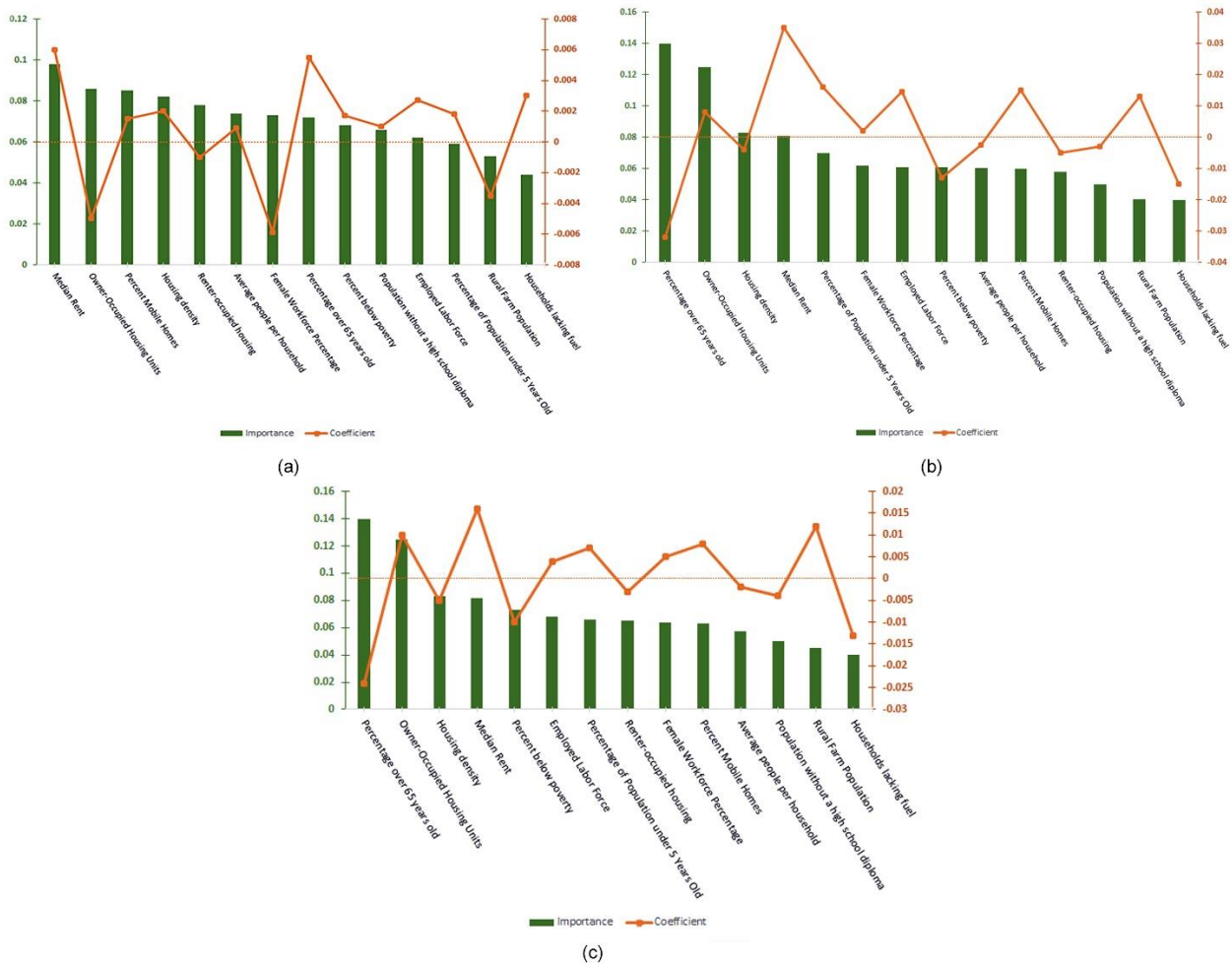

*Figure 5. Graphical relations of socioeconomic variables to the (a) vulnerability, (b) adaptability, and (c) resilience scores.*

For vulnerability modeling, the linear regression model suggests that variables with the highest coefficient magnitudes are the percentage over 65 years old, median rent, and percentage of female workforce. Among them, median rent, and percentage over 65 years old were positively contributing to vulnerability, while female workforce percentage impacted it negatively. On the other hand, the random forest model identified median rent, owner-occupied housing units, and percent mobile homes as the most important factors in shaping community vulnerability. Some notable actionable interpretations suggest owner-occupied housing units and female workforce percentage are significant in decreasing vulnerability, highlighting the role of stable homeownership and economic empowerment of women in reducing disaster vulnerability, thus enhancing community resilience. On the other hand, median rent positively affects vulnerability. While this may seem counterintuitive—as higher rent often signifies wealthier status—it could be attributed to the fact that expensive properties are often situated in risk-prone areas such as flood plains or coastal regions. The proximity to water bodies provides recreational value, but also a higher vulnerability.

For adaptability modeling, the linear regression model points to median rent, the percentage of the population under 5 years old, and the percentage of the population over 65 years old as variables with the highest coefficients. While the percentage over 65 years old negatively influences adaptability, the other two factors positively affect it. Conversely, the random forest model places the percentage of the population over 65 years old, owner-occupied housing units, and the housing density at the forefront of adaptability factors. The results can be interpreted as a younger population imply a higher potential for change and adaptation, with families being more motivated to protect their children. It also suggests that wealthier communities might have superior recovery strategies increasing adaptability, as well as a considerable challenge is posed by an aging population.

Lastly, in the case of resilience, the linear regression model highlights the percentage over 65 years old, median rent, and households lacking fuel as having the highest coefficients, with percentage over 65 years old and households lacking fuel negatively affecting resilience. The random forest model, aligning somewhat differently, emphasizes percentage of the population over 65 years old, owner-occupied housing units, and the housing density as the highest factors in influencing resilience. At most of these factors, while the magnitude of coefficients differs, it follows a similar trend to the adaptability relations. Thus, while economic resources and stable, owned housing contribute positively to overall resilience, the presence of a larger elderly population presents challenges. Additionally, some notable insights from other factors include a higher average "*number of people per household"* may strain resources and make it more difficult to secure adequate shelter, supplies, and evacuation options during disasters contributing negatively to resilience. Similarly, it is observed that larger households may face more difficulties during evacuation, with more people to account for and potentially move to safety.

However, these interpretations are based on associations and do not establish causality. To further investigate the potential causal relationships among the socioeconomic variables and their respective scores, a Bayesian network (Scutari, M. et al. 2010) model was employed using bootstrap subsampling. This case study used the Peter-Clark(PC)-stable algorithm (Tsagris, 2018), an enhanced version of the PC algorithm to learn the structure of a Bayesian network from observational data with improved stability and reliability. By iteratively and conditionally testing for independence between variable pairs, the algorithm constructs a skeleton of the Bayesian network, followed by edge orientation to infer causal relationships. The modification upon the original PC algorithm ensures all conditional independence tests for a given variable are completed before any edges are removed. Figure 6a, 6b and 6c depicts the Directed Acyclic Graphs derived from Bayesian Networks for vulnerability, adaptability, and resilience scores, respectively.

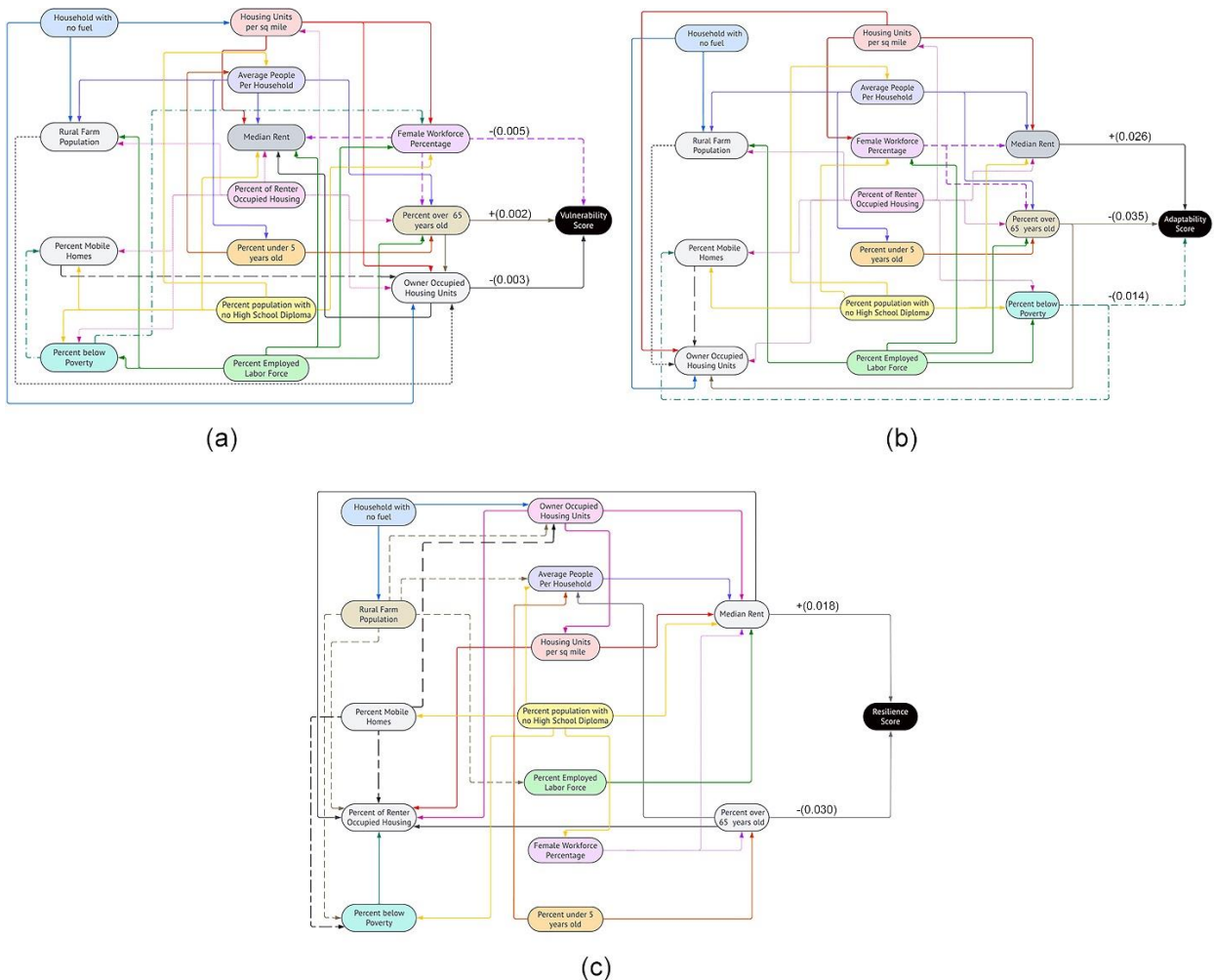

*Figure 6: Directed Acyclic Graphs of Socioeconomic variables to the (a) Vulnerability, (b) Adaptability, and (c) Resilience scores.*

This method can discover the causal relationships between the resilience scores and the socioeconomic factors, and the interplay between socioeconomic factors themselves. In this process, it finds the parent nodes of vulnerability, adaptability, and resilience as they share a direct causal relationship with a county's resilience scores. The analysis yielded 40 directed arcs (figure 6a) with *Female Workforce Percentage, Percent over 65 years old* and *Owner-Occupied Housing Units* emerging as the parent nodes to the Vulnerability score; 34 directed arcs (figure 6b) with *Percent over 65 years old, Median Rent* and *Percentage below poverty* emerging as the parent nodes to Adaptability score; and 32 directed arcs (figure 6c) with *Percent over 65 years old* and *Median Rent* emerging as the parent nodes to the resilience score. In the adaptability bayesian network*,* Percent over 65 years old and Percentage below poverty have negatively affected adaptability, while median rent has positively affected it. Subsequently, in the vulnerability model, Female Workforce Percentage and Owner-Occupied Housing Units negatively affect vulnerability while Percent over 65 years old positively affects it. Finally, the resilience Bayesian network has a positive coefficient for median rent and negative coefficient for Percent over 65 years old. The remaining directed arcs indicate causal relationships between the other variables, which potentially affect resilience scores indirectly.

## 6. Discussion and Conclusion

This paper showcases the establishment of a dedicated cyberinfrastructure platform, PRIME, specifically designed for resilience assessment and visualization. The platform's extensive pool of potential users, ranging from public officials, researchers, and the public, highlights its role in resilience assessment and risk management. The unique strength of the CyberGIS infrastructure is facilitating customizable disaster vulnerability, adaptability, and resilience assessments at county levels and identification of driving socioeconomic factors of these indices. By integrating CRIM into the CyberGISX infrastructure, we extend access to the datasets of hazard events, associated damages, and various resilience-related socioeconomic factors to a broader audience. Government entities can harness this information to aid decision-making processes in hazard planning. For the research community, it opens new horizons for fresh perspectives in resilience assessment, inspiring novel research directions. This work also provides a demonstration of how these models can be used and interpreted for nationwide analysis. Some of our notable outputs identify that the percentage of older population increases vulnerability as well as decreasing adaptability. Interestingly, owning houses has positively contributed to adaptability and reduced vulnerability. A contrasting result is seen in the case of younger population where they not only increase vulnerability of an area but also the adaptability.

However, several inherent limitations exist in this approach. Despite the CyberGIS infrastructure's significant strides in improving accessibility and comprehensibility of resilience-related data, its efficacy is contingent upon the availability and accuracy of the underlying datasets. Potential discrepancies or gaps in the data could result in assessment inaccuracies. Moreover, while the infrastructure enables identification of key resilience indicators, the interpretation and application of this information largely relies on the users' expertise.

Lastly, there may be concerns raised on the yearly calculation of these scores, given certain disasters like drought, expand for more than one year. We suggest considering the idea that these yearly scores, when combined, can provide an overall sense of a community's resilience over time. Essentially, we propose looking at the long-term resilience of a community as if it were the average of its annual resilience scores. For example, if a county has smaller recovery scores in the aftermath of Hurricane Harvey but recovers significantly at a later period, the average will show a high recovery score when calculated over a span of years. However, this definition of long-term indices is proposed purely for interpretation purposes and is not integrated into our calculations. It is a simplified perspective intended to provide an overarching view of the community's disaster resilience over a longer span, but it does not capture the nuances and fluctuations that occur from year to year.

In summary, the CyberGIS infrastructure PRIME represents a significant advancement in the quest for comprehensive, accessible resilience assessment tools. It offers a promising foundation for further refining our understanding of disaster resilience and enhancing our capacity for effective risk management. Possible future research employing this infrastructure could extend to comparative analysis, predictive analysis, etc. for varied county groups of interest.

**Data Availability Statement:**

The CyberGIS framework is available in here:
https://cybergisxhub.cigi.illinois.edu/notebook/prime-a-cybergis-platform-for-resilience-inference-measurement-and-enhancement/

**Acknowledgements**

This research is based in part on work supported by the U.S. National Science Foundation under grant numbers 2118329 and 2112356.